\documentclass{article}

    \usepackage[preprint]{neurips_2023}

\usepackage[utf8]{inputenc} % allow utf-8 input
\usepackage[T1]{fontenc}    % use 8-bit T1 fonts
\usepackage{hyperref}       % hyperlinks
\usepackage{url}            % simple URL typesetting
\usepackage{booktabs}       % professional-quality tables
\usepackage{amsfonts}       % blackboard math symbols
\usepackage{nicefrac}       % compact symbols for 1/2, etc.
\usepackage{microtype}      % microtypography
\usepackage{xcolor}         % colors

\usepackage{color}              
\usepackage{amsmath}

\DeclareMathOperator*{\argmax}{argmax}
\DeclareMathOperator*{\smax}{smax}

\DeclareMathOperator*{\explore}{explore}
\DeclareMathOperator*{\exploit}{exploit}

\usepackage{bbm}
\usepackage[suppress]{color-edits}
\addauthor{tl}{cyan}
\addauthor{bk}{red}
\addauthor{yx}{blue}
\addauthor{ja}{magenta}

\usepackage[lined,ruled,commentsnumbered]{algorithm2e}

\usepackage{graphicx}
\usepackage{bm}
\usepackage{amssymb, amsmath, graphicx, subfigure, enumerate}
\usepackage{amsthm,alltt} 
\usepackage{graphicx,ctable,booktabs}
\usepackage{mathtools}

\newcommand{\xhdr}[1]{\vspace{1mm} \noindent{\bf #1}}
\newtheorem{theorem}{Theorem}

\newtheorem{lemma}[theorem]{Lemma}

\newtheorem{definition}{Definition}
\title{On the Robustness of Epoch-Greedy in Multi-Agent Contextual Bandit Mechanisms}

\author{Yinglun Xu \\
  University of Illinois at Urbana-Champaign\\
  \texttt{yinglun6@illinois.edu} \\
  \And
  Bhuvesh Kumar \\
  Georgia Institute of Technology \\
  \texttt{bhuvesh@gatech.edu} \\
  \AND
  Jacob Abernethy \\
  Georgia Institute of Technology \\
  \texttt{prof@gatech.edu} \\
}

\begin{document}

\maketitle

\begin{abstract}
Efficient learning in multi-armed bandit mechanisms such as pay-per-click (PPC) auctions typically involves three challenges: 1) inducing truthful bidding behavior (incentives), 2) using personalization in the users (context), and 3) circumventing manipulations in click patterns (corruptions). Each of these challenges has been studied orthogonally in the literature; incentives have been addressed by a line of work on truthful multi-armed bandit mechanisms, context has been extensively tackled by contextual bandit algorithms, while corruptions have been discussed via a recent line of work on bandits with adversarial corruptions. Since these challenges co-exist, it is important to understand the robustness of each of these approaches in addressing the other challenges, provide algorithms that can handle all simultaneously, and highlight inherent limitations in this combination. In this work, we show that the most prominent contextual bandit algorithm, $\epsilon$-greedy can be extended to handle the challenges introduced by strategic arms in the contextual multi-arm bandit mechanism setting. We further show that $\epsilon$-greedy is inherently robust to adversarial data corruption attacks and achieves performance that degrades linearly with the amount of corruption.
\end{abstract}

\section{Introduction}

Auctions for allocating online advertisements are run many billions of times per day and constitute the main business models for many digital platforms. %these auctions generate a massive share of revenue for much of the industry.
How these auctions are designed and managed can therefore have major implications for both the auctioneers as well as the many many participants competing for ad space. Under a simple model of bidding behavior, and with good information about the quality of submitted ads, classical theory in mechanism design provides the auctioneer with what is essentially the uniquely optimal allocation and payment scheme. The VCG mechanism  \cite{vickrey1961counterspeculation} allocates each advertisement slot to the highest bidder, weighted by the ``click-through-rate'' (CTR), and charges an amount equal to the \emph{second highest} bid \jacomment{Cite some VCG papers!}. This scheme to select and charge bidders is welfare-maximizing and \emph{truthful}, i.e. such that the bidders do not have any incentive to misreport their true value for clicks.

The simple VCG strategy, while nice in theory, is unfortunately lacking in one critical way: it requires that the auctioneer possess knowledge of the click-through-rates of the various ads submitted by bidders. From this perspective, selecting an ad to display to the user is akin to pulling an arm in a bandit problem, where the welfare associated to a pull is value the bidder receives in the event of a click. Indeed, this $K$-armed bandit perspective on repeated ad auctions has been quite thoroughly studied; in particular, \citet{DBLP:journals/siamcomp/BabaioffSS14,devanur2009price} model each arm as an advertiser with an associated fixed click-through-rate. The key hurdle for applying bandit methods to repeated auctions is the conflict between minimizing regret and maintaining reasonable long-term incentives for the bidders, since the auctioneer's learning strategy may lead to unpredictable bid behavior. \citet{DBLP:journals/siamcomp/BabaioffSS14,devanur2009price} design a VCG style mechanism using an \emph{explore-then-commit} bandit strategy. The resulting mechanism is truthful and obtains the $O(T^{2/3})$ lower bound on the welfare regret which is measured with respect to the fixed arm with the highest expected welfare.

In this paper, we turn our attention to the \emph{contextual} version of the same bandit mechanism design problem.  In the above setting, while the auctioneer's estimates of the CTRs may change, the underlying CTRs for all the arms are fixed. But indeed these CTRs may not only be unknown but also fluctuate according to some \emph{context} that arrives before each auction and provides details about ad desireability (e.g. profile of the user, time of day, etc.). When the mechanism is given access to this context, as well as a family of experts (i.e. hypotheses, policies, etc.), it will want to minimize regret not with respect to simply the best arm (ad) but the optimal expert for selecting an arm according to any given context. This more general setting provides additional power to the mechanism designer, who can now consider much more complex policies for allocation that achieve higher welfare.  

Although contexts have been extensively studied in the multi arm bandits literature \cite{lu2010contextual,chu2011contextual,LangfordZhang07,AuerCeFrSc2003} \jacomment{add citations here}, there has been little focus on the interplay between contexts and strategic arms who care about their own long term utility. Giving more flexibility to the mechanism to select ads, in the form of contexts and policies, presents new challenges to the mechanism design problem as it can affect bidder incentives in ways that have not been previously addressed. Consider just this one example: a participant can bid higher for a particular ad slot, but this action can (all else equal) \emph{decrease} their likelihood of winning the slot, as the change can lead to the learning algorithm to move to a different policy. In order to manage this awkward incentive issue we slightly weaken our notion of truthfulness by considering a property we call $\alpha$-\emph{rationality},  where agents can only violate truthful bidding if it improves their utility by at least $\alpha$. Ultimately, we are able to show that the mechanism can guarantee regret on the order of $\tilde O\left( T^{2/3}K^{4/3} + T^{1/3}K^{2/3}/\alpha^2\right)$, with respect to the optimal arm selection policy, by using a relatively standard $\epsilon$-Greedy style algorithm. Pay per click auctions in the contextual multi-slot setting has also been studied before by \citet{gatti2012truthful} where the principal directly tries to learn the click through rate as a function of the contexts for each arm under strong assumptions about the class of functions it is learning over. Their results do not apply to our setting as we don't directly aim to learn the click through-rate for each context and arm but utilise the information given by the contexts through a class of experts.

Apart from the challenges induced by contexts and strategic agents, online ad auctions are also vulnerable to data corruption attacks such as click fraud \cite{wilbur2009click} \yxcomment{add citations}. For example, an agent may create a bot that either clicks her own ads or does not click competing ads to adversarially bias the click-through-rate estimates and make the ad seem less desirable to the principal. Corruption robustness in multi arm bandits has been studied at great length in recent years \cite{lykouris2018stochastic,gupta2019better} \jacomment{Please cite}, but never in multi arm bandit mechanisms. Although these algorithms have great performance in the stochastic bandit setting, they lack a key property called \emph{exploration separation} \cite{DBLP:journals/siamcomp/BabaioffSS14} which has been shown to be fundamental for ensuring truthfulness in bandit mechanisms. 

Finally, we show that the contextual $\epsilon$-greedy mechanism (Algorithm \ref{alg:contextmab}) has a surprisingly nice additional property: it is tolerant to a certain amount of adversarial click corruptions. In the easier case of stochastic multi arm bandit mechanism, we show that the contextual  $\epsilon$-greedy mechanism is truthful without any relaxation on the rationality of the agent and is inherently robust to adversarial click corruptions without any changes to the mechanism. If the number of rounds corrupted by the adversary is $C$, which is unknown to the auctioneer, the welfare regret of the mechanism in this setting is $\Tilde{O}(K^{1/3}T^{2/3}+C)$, which recovers the lower bound up to poly log terms. In the presence of contexts and $\alpha$-rational agents, the welfare regret of the contextual $\epsilon$-greedy mechanism is $\tilde O\left( T^{2/3}K^{4/3} + T^{1/3}K^{2/3}/\alpha^2 + C/\alpha + KC \right)$.

\paragraph{Other related work: }Our work contributes to the literature of principal learning in auction design in repeated settings with strategic agents. Outside of pay-per-click auctions there are multiple auction settings where principal learning with strategic agents arises, including posted prices \cite{amin2013learning,amin2014repeated,mohri2014optimal,mohri2015revenue}, reserve prices \cite{liu2018learning}, or revenue maximizing auctions \cite{abernethy2019learning}. These works assume either the bidders discount their future utilities or assume that the same bidders only shows up for a limited number of rounds and thus have limited impact over principal learning. We don't make such assumptions about the agents' behavior. Another line of work focuses on multi arm bandits in the presence of strategic arms \cite{nazerzadeh2016sell, braverman2017multi} which pass on part of their reward to the principal when picked.  One more related research direction involves mechanism design questions when the principal knows that the agents are employing no-regret learning to decide their actions  \cite{deng2019prior, heidari2016pricing}. The inverse problem of reacting as a buyer to a no-regret seller has also been studied in \cite{heidari2016pricing}. Finally, we note that repeated auction design with strategic agents also arises in many other works within dynamic mechanism design \cite{bergemann2006efficient, nazerzadeh2008dynamic, kakade2013optimal} for a non-exhaustive list. 

\section{Preliminaries}
\label{sec:model}
We study the problem of pay-per-click (or PPC) ad auctions in a multi-armed contextual bandit setting. There is a set of advertisers (also referred to as \emph{agents} or \emph{arms}) which we denote as $\mathcal{A}$ and has cardinality $K$; agents repeatedly compete for a single slot over $T$ time steps (that correspond to users). If an advertiser $a\in\mathcal{A}$ is selected, then a click occurs with probability equal to the click-through-rate. Upon a click, the agent earns value equal to $\mu(a) \in[0,1]$ (advertisers earn a fixed reward whenever a click happens, but different users may have different probability of clicking their ad which depends on the context of the user). Contexts come from a context space $\mathcal{X}$ and are identically and independently distributed (i.i.d.) across rounds. The click-through-rate function $\rho:\mathcal{A}\times\mathcal{X}\rightarrow [0,1]$ maps agents and contexts to the probability of being clicked upon display.

In each round $t = 1, \cdots, T$, a context $x^t \in \mathcal{X}$ is drawn i.i.d. from a fixed distribution. Each agent $a \in \mathcal{A}$ submits a bid $b^t(a)$ and, in response the platform selects an agent $a^t$ and displays her ad. The selected ad is clicked with probability $\rho(a^t,x^t)$ and the corresponding clicked result $c^t$ is $1$ if the ad is clicked and $0$ otherwise. The platform receives the click result $c^t$ and if the ad gets clicked, i.e. $c^t = 1$, then it charges the agent $a^t$ a payment $p^t$ and the agent $a^t$ gains utility $u^t(a^t) = \mu(a^t) - p^t$. If the ad is not clicked, $c^t = 0$, then the payment and the utility of the agent is $0$.

Following classical works on contextual bandits, we also assume a class of hypotheses (also referred to as experts) which we denote as $\mathcal{H}$ with $|\mathcal{H}| = m$. Each expert $h: \mathcal{X} \rightarrow \mathcal{A}$ is a mapping from the context space to agents where $h(x)$ is the agent recommended by the expert $h$ given context $x$. The auctioneer can use the experts' recommendations to make decisions.

\xhdr{Objective.} The platform wishes to maximize the cumulative social welfare defined as $\sum_t c^t \cdot \mu(a^t)$. Note that $c^t$ is a random variable that depends on (possibly randomized selection of) agent $a^t$ and on the randomness in the context $x^t$. To evaluate how well the platform performs, we need a benchmark and the typical benchmark is to compare against the expected social welfare of the best expert $h\in\mathcal{H}$ where the expectation is taken over the randomness in the contexts. When the value profile of the agents is $\vec{\mu}\in [0,1]^K$ where $\mu(a)$ is the value of agent $a$, the expected welfare for an expert $h$ is $R(h,\vec{\mu}) = \mathbbm{E}_{x}[\mu(h(x))\cdot \rho(h(x),x)]$. Note that $\rho(h(x),x)$ is the expected value of the click result when the context is $x$ and the arm suggested by the expert $h$, i.e. $h(x)$ is displayed.

Let $h^\star=\argmax_{h \in \mathcal{H}}\{R(h,\vec{\mu})\}$ be the expert with the highest expected welfare. The performance of our algorithm against $h^\star$ is measured by the notion of regret:

$$\textsc{Reg}_\text{context} = T\cdot R(h^\star,\vec{\mu}) - \sum_{t=1}^T c^t\cdot \mu(a^t).$$

If we assume that agents report their value truthfully (defined below) then this reduces to the contextual bandit setting \cite{AuerCeFrSc2003,LangfordZhang07}. 

A special case of this setting is the multi-armed bandit setting where the hypothesis class is the same as the agent space $\mathcal{H}=\mathcal{A}$ and each hypothesis always suggests the same agent. In this special case the expected welfare of hypothesis $h=a$ is $R(h,\vec{\mu}) = \mu(a)\cdot \mathbbm{E}_{x}[\rho(a,x)]$.

%If the principal has access to the value profile and expected click through rates for each arm, then the welfare optimizing option is to always select the arm with the highest expected welfare $i^* = \argmax_i \bar{\rho}_i \cdot \mu_i$. The corresponding benchmark is $$\textsc{StoOpt}(T) = T \cdot \bar{\rho}_{i^*}\cdot \mu_{i^*}.$$

%Let the arm that selected by the principal at round $t$ as $a^t$. The welfare regret for a bandit algorithm is defined as 

\xhdr{Incentives.}  \citet{DBLP:journals/siamcomp/BabaioffSS14} and \citet{devanur2009price} introduced incentives in the multi-armed bandit setting by considering that agents bid strategically with the goal to maximize their own long term utility defined as the difference between the total welfare they collect and the price they pay over the duration of auction. They focus on designing mechanisms that are \emph{truthful}, i.e., agents cannot benefit by bidding something other than their value. They show that the strategic behaviour of agents poses additional challenges by providing a $\Omega(T^{2/3})$ lower bound for this setting compared to the $O(\sqrt{T})$ regret that is achievable in the absence of incentives. This $O(T^{2/3})$ regret can be achieved by classical algorithms that separate exploration from exploitation such as explore-then-commit and $\epsilon$-greedy.

In this work, we introduce the study of truthful mechanism design in contextual bandit mechanisms. Our definition of truthfulness generalizes the one in multi-armed bandits:

\begin{definition}[Truthful Mechanism]
\label{def:thruthful}
A mechanism is \emph{truthful} if for any click-through-rates function $\rho(\cdot,\cdot)$ and value profile $\vec{\mu}$, for every agent $a \in \mathcal{A}$, irrespective of the bidding strategies used by other agents ($\mathcal{A} \setminus a$), agent $a$ obtains the maximum expected utility over all $T$ rounds by bidding her true value $b^t(a)=\mu(a)$ at every round $t$.
\end{definition}

We now introduce some definitions needed to characterize the setting. Suppose the expert that the mechanism follows is $h$, and the context is $x$, then the probability that agent $a$ gets a click is its click probability under context $x$ times the indicator of whether $h$ suggests agent $a$, i.e., $\rho(a,x)\mathbbm{1}\{h(x)=a\}$. Since the context in each round is identically and independently distributed, the probability that the agent will receive a click is this value in expectation over all contexts, which gives the following definition.

For the rest of the paper, we assume that the agents cannot observe the context. This is necessary as we show in the appendix \ref{sec:observe}\bkcomment{cite appendix} that if agents are allowed to observe the context, then it is hard to design truthful and low-regret mechanism without any assumptions on the expert class. This is a justifiable assumption as the context may contain sensitive information about the users which should not be shared with the advertiser. 

\begin{definition}[Click probability for an agent induced by an expert]

The click probability for an agent $a$ induced by an expert $h$ is the probability that the agent will receive a click if the principal follows the recommendations of expert $h$, which is given by $$\Pi(a,h) := \mathbbm{E}_x[\rho(a,x)\mathbbm{1}\{h(x)=a\}],$$
where $p(x)$ is the probability that the context is $x$. 
\end{definition}

 Note that based on the definition of $\Pi(a,h)$, the expected welfare of the mechanism following an expert $h$ can be written as $ R(h,\bm{\mu}) = \sum_{a \in \mathcal{A}} \mu(a)\cdot \Pi(a,h)$. Next we introduce the definition of reported welfare which is a proxy for expected welfare.

\begin{definition}[Reported welfare of an expert]
The reported welfare of an expert $h$ when the bids profile is $\bm{b}$ is:

$$ R(h,\bm{b}) = \sum_{a \in \mathcal{A}} b(a)\cdot \Pi(a,h)$$
\end{definition}

Note that when the mechanism is truthful, agents will bid their true values, i.e. $\bm{b}=\bm{\mu}$, then the reported welfare of an expert will be identical to its expected welfare. In this case, the expert with the highest reported welfare is exactly the best expert, that is, $\argmax_{h \in \mathcal{H}}R(h,\bm{b}) = h^*$. 

Next we want to characterize the incentives of agents. One important quantity the incentives are based on is the relation between how much they bid and how likely they will get a click. Recall that the mechanism allocates slots to agents based on experts class $\mathcal{H}$, data in the history $H^t$ which includes all the information available to the principal up to round $t$, agents' bids $\bm{b}$, and the context $x$. So for an agent $a$, the probability that it could win the slot if bidding $b(a) = b$ is $\Pr\{A^t(\mathcal{H}, H^t, x, b, \bm{b}^t_{-a})=a\}$, where $A^t$ is the allocation rule of the mechanism at round $t$ and $\bm{b}^t_{-a}$ is the bids of the other agents, which multiplied by the click through rate $\rho(a,x)$ would be the probability that it could receive a click. Since the agent cannot observe the context, the general probability to receive a click should be that value in expectation over all context, which gives the following definition.

\begin{definition}[Click probability of an agent]
\label{def:g}
Let $A^t$ be the allocation rule used by the mechanism at round $t$, $H^t$ be the history till round $t$, and $\mathcal{H}$ be the experts class. For an agent $a$, fixing the bids of the other agents as $\bm{b}^t_{-a}$, then the probability that the agent $a$ get clicked in round $t$ by bidding $b(a)=b$ is given by 

 $$g^t(a,b) = \mathbbm{E}_{x}[\Pr\{A^t(\mathcal{H}, H^t, x, b, \bm{b}^t_{-a})=a\}\rho(a,x)]$$ 
 
 where $\rho(a,x)$ is the click through rate for agent $a$ on context $x$.
\end{definition}

Based on the definition of $g^t(a,b)$, we are able to formally write the expected utility function of an agent $a$ when her bid is $b$ as:$$u^t(a,b)=(\mu(a)-p^t(a,b))g^t(a,b),$$ where
$p^t(a,b)$ is the payment at round $t$ if agent $a$ is selected and clicked and her bid is $b$. Note that the function $p^t(\cdot,\cdot)$ can depend on the allocation rule, bids from other arms, data of history, and experts class. 

Finally, we introduce the definition of exploration separated allocation rule which has been shown to be fundamental for truthful bandit mechanisms \cite{DBLP:journals/siamcomp/BabaioffSS14}.

\begin{definition}[Exploration-separated allocation rule (Definition 1.2, \citet{DBLP:journals/siamcomp/BabaioffSS14})]
An allocation rule is called \textbf{exploration-separated} if the allocation on any influential rounds does not depend on any bids, where influential rounds are the rounds whose click result and bids will influence the allocation in the future rounds.
\end{definition}

\section{Contextual bandit mechanisms}
\label{sec:context}
%Unlike the multi-armed bandit setting, an interesting question that arises has to do with the information that is available to the agent when making a bid. As we show in Section~\ref{sec:negative}, if agents can bid based on the context then linear regret is unavoidable. This is because by their bid, agents

% If contexts are indeed present and each experts predictions might change with the context, the requirement that agents are not allowed to observe the context becomes crucial to maintain truthfulness. In section 3, we show that with such requirement, the $\epsilon$-greedy mechanism can effectively utilize the contexts through experts while maintaining approximate truthfulness. In section 4, we show that when agents can see the context, without any assumptions on the expert class, it becomes harder to achieve truthfulness.

%In this section, we require that the context is only available to the experts and cannot be observed by the agents. We show in section \ref{sec:negative} why such an assumption is necessary. 

In the stochastic setting without contexts, since each expert is an arm and we are competing with the best arm in expectation, as shown in \cite{DBLP:journals/siamcomp/BabaioffSS14, kakade2013optimal}, a simple VCG style mechanism recovers the lower bounds for the regret. In the presence of a more complex class of experts which can utilize the context to make better decisions, the benchmark becomes harder as we are now competing with the best expert in expectation. To compete with the best expert, the platform has to consider expert predictions while making decisions and the interaction between the agents and the experts makes truthfulness harder. 

We focus our attention on designing mechanisms using exploration separated allocation rules. For such allocation rules, in any non-influential round, i.e. the round whose click results and bids don't influence the future decisions of the principal, the bid reported by an agent can only influence the utility obtain in the current round. Hence the agents optimize their bid only to maximise the utility from the current non-influential round. In other words, for each $a$ and $t$ that is non-influential, if $u^t(a,b)$ is maximized at $b=\mu(a)$, then all agent will always bid truthfully in every non-influential round. This fact gives a necessary condition for a mechanism which uses an exploration separated allocation rule to be truthful.

\begin{lemma}\label{lem:payment}
If a mechanism uses an exploration separated allocation rule, then the mechanism is truthful only if for every non-influential round $t$ and every agent $a$, the probability that she gets clicked in round $t$, i.e. $g^t(a,b)$ is monotonically increasing in her bid $b$ and the payment rule satisfies $$p^t(a,b) \cdot g^t(a,b) = b \cdot g^t(a,b) - \int_{y=0}^{b}g^t(a,y)\mathrm{d}y$$
\end{lemma}

In fact, this payment rule can be viewed as the Pay-Per-Click auction version of the Myerson payment identity \citep{myerson1981optimal}. This theorem shows that for any mechanism using an exploration separated allocation rule, it can only be truthful for some payment rule only if $g^t(a,b)$ is always monotonically increasing in $b$ in any non-influential round $t$. 

Note that to have monotone $g^t(a,b)$, one simple allocation rule could be randomly picking an arm at each round, but clearly such allocation rule is far worse then the benchmark we want to achieve. We need a more complicated allocation rule which can guarantee monotonicity of $g^t(a,b)$ while recovering the benchmark. 

\paragraph{Perfect information.} If the click probability for an agent induced by an expert $\Pi(a,h)$ for all $a \in \mathcal{A}$ and $h \in \mathcal{H}$ is known by the principal at the beginning, to recover the benchmark, a straight forward idea is to always follow the expert with the highest reported welfare, i.e. $h^t = \argmax_h\{R(h,\bm{b})\}$ at each round. Mechanisms with such allocation rule will recover the benchmark if all agents bid truthfully, and the lemma below shows that there exist payment rules to make the mechanism truthful. 

\begin{lemma}\label{lem:optimaltruthful}
In contextual PPC auction, there exist payment rules such that the mechanism which always follows the expert with the highest reported welfare $h^t = \argmax_h R(h,\bm{b}^t)$ is truthful.
\end{lemma}

The intuition behind the proof is to show that the click probability of each agent $a$ as a function of her bid is monotonically increasing. This is true because if an agent increases her bid, and the expert selected by the mechanism changes, then the the click probability for that agent induced by the new expert can only be higher. 

Aside from the requirement of a monotone $g^t(a,b)$, the payment rule poses an additional difficulty in the contextual setting. In the easier case where experts are arms themselves as shown in \cite{devanur2009price}, the allocation rule that maximizes the welfare results in a simple $g^t(a,b)$ whose value is the click through rate of agent $a$ times the indicator of $b > d$ where $d$ is the weighted highest bid of the other bidders. The corresponding truthful payment rule given by lemma \ref{lem:payment} is $p^t(a,b) = p$ which doesn't require the knowledge of the click through rate. In the setting with more complicated class of experts, the allocation rule that chooses the expert with the highest reported welfare can result in much more complicated $g^t(a,b)$, and to find the corresponding truthful payment rule requires the full information of $g^t(a,b)$, which in turn requires the information of $\Pi(a,h)$ for all $a$ and $h$. If the payment rule a mechanism actually uses is only slightly deviated from the truthful payment rule, the utility from bidding truthfully is still close to the optimal, but the optimal bid can be far from their true value. We give an example in the appendix \ref{sec:example} to illustrate this fact. Due to this reason, we relax the requirement of truthfulness by introducing the definition of $\alpha$-rational agents, who will bid truthfully as long as bidding truthfully leads to utility close to the optimal utility.

\begin{definition}[$\alpha$-rational]
In mechanisms where the bids of agents of a round will not affect the future rounds, an $\alpha$-rational agent will bid truthfully in round $t$ if the extra utility in that round she can get by misreporting her value is bounded by $\alpha$. That is, an $\alpha$-rational agent $a$ will bid truthfully in round $t$ if $\max_b u^t(a,b) - u^t(a,\mu(a)) \leq \alpha.$
Note that $\alpha \in [0,1]$.
\end{definition}

Since estimating the optimal bids for the agents requires them to know information about the competition and other estimates of the mechanism, it is reasonable to assume that if the extra utility they get from trying to estimate their optimal bids is very small, then they can just bid truthfully.

\subsection{$\epsilon$-Greedy for Contextual Bandit Mechanisms}
Now we present the contextual $\epsilon$-Greedy Mechanism ( algorithm \ref{alg:contextmab}) for the contextual multi-arm bandit PPC auction setting. In each round, with a fixed probability $\epsilon$, the mechanism uniformly at random select an arm. Such rounds are called \emph{explore rounds} and the mechanism learns over each expert based on the information only from the explore rounds. For each arm $a \in \mathcal{A}$ and expert $h$, the mechanism tries to learn $\Pi(a,h)$ the click probabilities of agent $a$ induced by expert $h$ as empirical estimate $\hat{\Pi}(a,h)$ from the explore rounds. There is no payment charged in the explore rounds. The other rounds are called \emph{exploit rounds} where the principal follows the empirically best expert under the current bids and then selects the arm suggested by this expert. In case of a click, the mechanism also the estimates $\hat{\Pi}(a,h)$ to charge payments given by lemma \ref{lem:payment} with the true click probabilities of agents substituted by the empirical estimation.

First we show that contextual $\epsilon$-greedy mechanism can maintain accurate empirical estimates of the click probability for agents induced by experts.

\begin{lemma}\label{lem:error}
For the contextual $\epsilon$-greedy mechanism (algorithm \ref{alg:contextmab}), let the number of experts in $\mathcal{H}$ be $m$, number of rounds be $T$, and number of agents be $K$. With probability at least $1-3/T$, for every expert $h$, every arm $a$, at every round $t \geq 24\log T/\epsilon$ the estimation error on click probability is bound by $$|\hat{\Pi}^t(a,h)-\Pi(a,h)| \leq K\sqrt{\frac{2\log (T\cdot m\cdot K)}{t\epsilon}}.$$
\end{lemma}
Since in each round, we explore each arm with a fixed probability $\epsilon/K$, thus for each expert we get a sample with this probability, and the result follows from a simple concentration bound.

Since we use empirical estimates to calculate the payments, in the initial few rounds, agents might be able to take advantage of the inaccurate estimates, but after a few rounds, as the estimates become more accurate, we show that the the extra utility agents can get by misreporting is bounded. Thus if the agents are $\alpha$-rational, after a few rounds, the agents will start bidding truthfully. 

\begin{theorem}\label{thm:rationalguarantee}
For the contextual $\epsilon$-greedy mechanism (algorithm \ref{alg:contextmab}), let the number of experts in $\mathcal{H}$ be $m$, number of rounds be $T$, and number of agents be $K$. With probability at least $1-3/T$, an $\alpha$-rational agent will bid truthfully in all $t \geq \frac{8K^2 \log(T\cdot m\cdot K)}{\epsilon \alpha^2}$ rounds.
\end{theorem}

The intuition behind Theorem \ref{thm:rationalguarantee} is that when the estimates $\hat{\Pi}(a,h)$ are more accurate for all arms $a \in \mathcal{A}$ and experts $h \in \mathcal{H}$, and the corresponding payment rule is more accurate, thus the extra utility an agent can obtain from trying to exploit the errors in the estimate goes down as the rounds go by.

Using lemma \ref{lem:error} and theorem \ref{thm:rationalguarantee}, we now present the main regret guarantee for this setting. We show that the contextual $\epsilon$-greedy mechanism (Algorithm \ref{alg:contextmab}) has diminishing regret compare to the best expert when all the agents are $\alpha$-rational.

\begin{theorem}\label{thm:contextregret}
For the contextual $\epsilon$-greedy mechanism (algorithm \ref{alg:contextmab}), let the number of experts in $\mathcal{H}$ be $m$, number of rounds be $T$, and number of agents be $K$. If all agents are $\alpha$-rational, the expected welfare regret of contextual $\epsilon$-Greedy Mechanism satisfies

\begin{align*}
  \textsc{Reg}(T) \leq &T\epsilon +  K^2 \sqrt{\frac{8T\log(T\cdot m\cdot K)}{\epsilon}} + 3K +\\
  & \frac{8K^2 \log(T\cdot m\cdot K)}{\epsilon \alpha^2} 
\end{align*}

Setting $\epsilon = 2T^{-1/3}\log (T\cdot m\cdot K)^{1/3}K^{4/3}$, we have 

\begin{align*}
    \textsc{Reg}(T) \leq & \widetilde{O} \left( T^{2/3}K^{4/3} + T^{1/3}K^{2/3}/\alpha^2\right)
\end{align*}

\end{theorem}

The results come from the observation that regret can be decomposed into four parts: (a) explore rounds; (b) rare cases where estimation are not accurate at all; (c) the early few rounds when $\alpha$ rational agents do not bid truthfully; (d) exploit rounds when the estimates are accurate and agents are bidding truthfully. a) and b) can be bounded as a direct consequence of concentration inequalities, and c) is bounded using theorem \ref{thm:rationalguarantee}. For d), we show that even if we follow the suggestions of a sub-optimal expert, its expected welfare can not be too far from that of the best expert.

{
\begin{algorithm}
    
  %\SetKwInOut{Input}{Input}
  \SetKwInOut{Param}{Parameters}
  \SetKwInOut{Init}{Initialize}

   \SetAlgoLined 
   \DontPrintSemicolon
 
%   \myalg{}{

  \Param{Number of arms $K$, Number of rounds $T$, exploration rate $\epsilon$, A hypotheses class $H$}
  \Init{$\hat{h} \leftarrow $ an arbitrary expert in $H$, $t_{e} \leftarrow \{\}$ \tcc*[r]{$\hat{h}$ is the empirical best expert, $t_{e}$ is the index of explore rounds}
  }

  \For{$t=1, \ldots, T$}{
  
  Receive bids $b^t_j$ from arm $j$ for all $j \in [K]$

$\ell^t \leftarrow \begin{cases} 
      1 & \textnormal{w.p. } \epsilon \\
      0 & \textnormal{otherwise} \\
   \end{cases}$    \tcc*[r]{Explore or exploit}

  \eIf{$\ell^t = 1$}{
  \fbox{\tcc{\textbf{... Explore Round ...}}}
  
    $t_e \leftarrow t_e \cap \{t\}$
    
    $a^t \leftarrow j $ uniformly at random for $j \in [K]$  \tcc*[r]{Select arm}
    
    Receive click result $c^t$
    
    Charge $p^t \leftarrow 0$  \tcc*[r]{Payment}
  }{
  \fbox{\tcc{\textbf{... Exploit Round ...}}}
    
    \For{$h \in H$}{
    \For{$i \in [K]$}{
        $\hat{\Pi}^t(i,h) \leftarrow \sum_{\tau \in t_e} \mathbbm{1}(h(x)=i)\mathbbm{1}(a^t=i)c^\tau \cdot K/|t_e|$
        }
    }
    
    $h^* \leftarrow \argmax_h \sum_{i \in [K]} \hat{\Pi}^t(i,h)b^t_i$ \tcc*[r]{update the empirical optimal best expert}
    
    $h^*$ Receives context $x^t$ \tcc*[r]{only the experts have access to the context}
    
    $a^t \leftarrow h^*(x^t)$ \tcc*[r]{Select arm}
    
    Receive click result $c^t$
    
    \For{$y \in [0,1]$}{
    $b^t_{a^t} \leftarrow y$
    
    $h \leftarrow \argmax_{h \in H} \sum_{i \in [K]} \hat{\Pi}^t(i,h)b^t_i$
    
    $\hat{g}^t(a^t,y) \leftarrow \hat{\Pi}^t(a^t,h)$
    }
    Recover value of $b^t_{a^t}$
    
    Charge price $p_i^t \leftarrow \begin{cases} 
       b^t_{i} - \frac{\int_0^ {b^t_{i}} \hat{g}^t(i,z) \mathrm{d}z}{\hat{g}^t(i, b^t_{i})} & \textnormal{\textbf{if} } c^t = 1, a^t = i, \hat{g}^t(i,b^t_{i})>0 \\
      0 & \textnormal{otherwise} \\
   \end{cases}$ \tcc*[r]{Payment}
 }
}
\caption{Contextual $\epsilon$-greedy Mechanism} \label{alg:contextmab}
\end{algorithm} 
}
\section{ Robustness Against Data Corruption}
\label{sec:corruption}
In this section, we show that, in addition, $\epsilon$-greedy mechanism is intrinsically robust against data corruption. 
%The intuition is that $\epsilon$-greedy algorithm uses a fixed explore rate to collect data to form estimates, so the impact of the corruption on such estimates is limited since most corruption will not be observed, which makes the algorithm robust to the adversarial data corruption attack eventually. 
Before we present the results, we first give a general introduction of adversarial data corruption setting.

\xhdr{Adversarial corruptions.} An extension of the multi-armed bandit setting allows for the click results in some rounds to be corrupted; the click results in non-corrupted rounds are stochastic generated based on the click through rate, while in the corrupted ones, the click results is generated by an adversary instead.
Following the model of \citet{lykouris2018stochastic}, the adversary can observe all the history of past outcomes until round $t$ as well as the principal's distribution at round $t$ but does not have access to the random selection of arm $a^t$; this is consistent to the adversarial bandit literature. The adversary has a corruption budget $C$ which we term \emph{corruption level} and captures the number of rounds that the adversary is allowed to corrupt. The corruption level is unknown to the principal.

\subsection{Warm up: corruption robustness without contexts}
As a warm up, we start with the stochastic setting where each expert is an arm itself. As discussed earlier, \citet{DBLP:journals/siamcomp/BabaioffSS14} show that in the absence of corruptions, an explore then commit approach can recover the $O(T^{2/3})$ lower bound of this setting. However when there is adversarial corruption, the simplest explore then commit mechanism is not robust against data corruption. Since the result from the explore phase determines the decisions in the exploit phase, if the adversary is able to corrupt the initial explore phase of the mechanism, then the mechanism will make decisions based on the corrupted data where a sub-optimal arm may appears to be the optimal. The mechanism will always end up selecting this sub-optimal arm and suffer from linear regret.

The stochastic $\epsilon$-greedy mechanism (Algorithm \ref{alg:robustmab}) presented in Appendix is basically a randomized version of the explore then commit mechanism. Similar to the Contextual $\epsilon$-greedy mechanism (Algorithm \ref{alg:contextmab}) introduced in the contextual setting, in each round, with a fixed probability $\epsilon$, the mechanism selects the arm uniformly at random, and these rounds are called \emph{explore rounds}. Since each expert $h$ corresponds to a fixed arm, the click probability of agent $a$ induced by expert $h$ $\Pi(a,h)$ is $\rho(a) = \mathbbm{E}_x[\rho(a,x)]$ for $a=h$ and $0$ for the other arms. Thus, we only need to estimate $\rho(a)$ which is the click through rate of each arm in this case, which is done based on the results from the explore rounds only. The other others rounds are called \emph{exploit rounds} where the mechanism selects the empirically optimal arm given the current bids. The payment rule in the exploit rounds is a weighted second price rule, and the payment is always $0$ in explore rounds. 

In the stochastic $\epsilon$-greedy mechanism, the bids reported in any round have no influence over the other rounds. In explore rounds, the arms are chosen uniformly at random independent of the bid and the payment is $0$, hence truthful bidding leads to optimal utility of that round. In exploit rounds, the allocation and payment is given by a weighted second price auction, which is truthful for any weights. Thus, the stochastic $\epsilon$-greedy mechanism is truthful.

Unlike the deterministic explore then commit mechanism, the stochastic $\epsilon$-greedy is able to maintain an accurate estimation on arms' click through rate even in the existence of data corruption. This allows the mechanism to be robust to adversarial corruptions at any unknown level.

\begin{theorem}\label{thm:stochasticbound}
If the corruption level for the mechanism is C, let $C'=C+\max\{C,6\log (T) K/\epsilon\}$. The expected welfare regret of stochastic $\epsilon$-greedy mechanism satisfies
\begin{align*}
\textsc{Reg}(T) &\leq T\epsilon + \sqrt{\frac{8KT\log T}{\epsilon}} + \\
&\frac{24K\log T}{\epsilon} + 2(K+1) + 4C'\log T
\end{align*}

By setting $\epsilon = 2K^{1/3}(T\log T)^{-1/3}$, the welfare regret is bounded by $\textsc{Reg}(T) = \widetilde{O}(K^{1/3}T^{2/3}+C)$.
\end{theorem}

Similar to theorem \ref{thm:contextregret}, the total regret can be decomposed into three parts: (a) regret from explore rounds; (b) regret in the rare case when the estimation is not accurate; (c) regret from exploit rounds when the estimation is accurate. Again a) and b) are direct consequence of concentration bounds. For c), we show that when the estimation is accurate, the expected welfare from the selected arm will not be far from that of the best arm.

We have shown that stochastic $\epsilon$-greedy mechanism is simultaneously truthful and robust to the adversarial attack in the stochastic setting. Next we will show the contextual $\epsilon$-greedy mechanism is also robust against data corruption in the more general case of contextual setting under the assumption that the agents are $\alpha$-rational.

\subsection{Corruption in Contextual Setting}
Similar to stochastic $\epsilon$-greedy mechanism, the contextual $\epsilon$-greedy mechanism \ref{alg:contextmab}c is also robust to adversarial data corruption attacks as it is able to maintain accurate estimates on click probability of agent $a$ induced by expert $h$, $\Pi(a,h)$ even in the presence of corruptions.

\begin{theorem}\label{thm:ccontextregret}
Let the corruption level be $C$ and denote $C' = C + \max\{C,\frac{6\log T}{\epsilon}\}$. Let the number of experts in $\mathcal{H}$ be $m$. If all agents are $\alpha$-rational, the expected welfare regret of contextual $\epsilon$-Greedy Mechanism is bound by 

\begin{align*}
  \textsc{Reg}(T) \leq &T\epsilon +  K^2 \sqrt{\frac{8T\log(T\cdot m\cdot K)}{\epsilon}} + 3K +\\
  & \frac{8K^2 \log(T\cdot m\cdot K)}{\epsilon \alpha^2} + \frac{16C'}{\alpha}+4KC'\log T.
\end{align*}

By taking $\epsilon = 2T^{-1/3}\log (T\cdot m\cdot K)^{1/3}K^{4/3}$, we have $\textsc{Reg}(T)\leq \widetilde{O}(T^{2/3}K^{4/3}+T^{1/3}K^{2/3}/\alpha^2+C'/\alpha+KC')$

\end{theorem}

\paragraph{Conclusion.} Recall that the three challenges of efficient learning in multi-armed bandit mechanisms as aforementioned include: 1) inducing truthful bidding behavior (incentives), 2) using personalization in the users (context), and 3) circumventing manipulations in click patterns (corruptions). We have shown that $\epsilon$-greedy mechanism can guarantee low regret in the contextual setting while maintaining a weaker notion of truthfulness under the relaxation that agents are $\alpha$-truthful, and it is also intrinsically robust to adversarial data corruption attacks.

\bibliographystyle{plainnat}
\bibliography{references}

\newpage

\onecolumn
\appendix
\section{Appendix}
\subsection{Proofs}
\paragraph{Proof for lemma \ref{lem:optimaltruthful}}
If a round is an explore rounds, then the bid of an agent neither influence the allocation or payment of the current round, nor influence the future, so bidding truthfully can lead to the highest utility in this case. Next we show that there exists payment rule such that bidding truthfully will also lead to the highest utility in exploit rounds. By lemma \ref{lem:payment}, we only need to prove that for this mechanism, $g(a,b)$ is monotonically increasing in $b$ for any $a$. Suppose the best expert is $h^*_1$ when the bid profile is $\bm{b}$, and $h^*_2$ when the bid profile is $\bm{b}+\Delta \cdot \bm{e}(a)$ where $\Delta > 0$ represents an increase in agent $a$'s bid. We need to show that $\Pi(a,h^*_2)\geq \Pi(a,h^*_1)$.

By definition of best expert given a bid profile, we have $$\sum_{i \in \mathcal{A}} b(i) \Pi(i,h^*_1) \geq \sum_{i \in \mathcal{A}} b(i) \Pi(i.h^*_2) ,$$ and $$\sum_{i \in \mathcal{A}} (b(i)+\mathbbm{1}\{i=a\}\Delta) \Pi(a,h^*_2)\geq \sum_{i \in \mathcal{A}} (b(i)+\mathbbm{1}\{i=a\}\Delta) \Pi(a,h^*_1).$$

Adding up both gives $\Delta(\Pi(a,h^*_2)-\Pi(a,h^*_1))\geq 0$, which means $\Pi(a,h^*_2)\geq \Pi(a,h^*_1)$. The proof is concluded.

\paragraph{Proof for theorem \ref{thm:stochasticbound}}

First we give the guarantee on the number rounds that are decided to be explore rounds and the amount corruption that are received in explore rounds. Denote $n^t(a)$ as the number of times when arm $a$ get selected in explore rounds. The expectation of $n^t(a)$ is $\mathbbm{E}[n^t(a)]=t\epsilon/K$. By Chernoff lower tail given in Lemma \ref{lem:Chernoff}, taking $\delta=1/2$, then we have $\Pr\{n^t_j<\frac{t\epsilon}{2K}\}\leq 1/T^2$ when $t\geq 24K\log T/\epsilon$. Denote $C^t(a)$ as the amount of corruption that arm $a$ received when it get selected in explore rounds by round $t$. The expectation of $C^t(a)$ is no greater than $\frac{\epsilon C}{K}$. If $C<6K\log T/\epsilon$, by Chernoff lower tail, taking $\delta = \frac{6K\log T}{C \epsilon}$, we have $\Pr\{C^t(a) > \frac{\epsilon}{K}(C+\frac{6K\log T}{\epsilon})\} \leq 1/T^2$; if $C>6K\log T/\epsilon$, take $\delta = \sqrt{\frac{6K\log T}{C\epsilon}}$, we have $\Pr\{C^t(a) >\frac{2\epsilon C}{K}\}\leq 1/T^2$.

Next we give the guarantee on the accuracy of the estimate on click through rates of arms. Combing above, when $t\geq 24K\log T/\epsilon$, with probability at least $1-2K/T$, we have $n^t(a)>\frac{t\epsilon}{2K}$ and $C^t(a)<(C+\max\{C,\frac{6K\log T}{\epsilon}\})\frac{K}{\epsilon}$ for all $j \in [K]$ and $t \in [T]$. Denote $C'=C+\max\{C,\frac{6K\log T}{\epsilon}\}$. Note that the deviation from estimate on arms' click through rate to the real value is given by the nature of stochastic which can be bound by concentration inequality, and by the adversarial corruption which is bound by $C^t(a)$. Then we have $\Pr\{|\hat{\rho}^t(a)-\rho(a)|<\sqrt{\frac{\log T}{\frac{t\epsilon}{2K}}}+\frac{2C'}{t}\} \geq 1-2(K+1)/T$ for all $a \in \mathcal{A}$ and $t\geq 24K\log T/\epsilon$.

Finally we can bound the regret. Let's divide the $\textsc{Reg}(T)$ into the regret from exploration rounds $\textsc{Reg}_{\explore}(T)$ and the regret from exploitation rounds $\textsc{Reg}_{\exploit}(T)$.

The expected number of explore rounds is $T\epsilon$. Thus $\textsc{Reg}_{\explore}(T) \leq T\epsilon$.

Let $w^t = \sqrt{\frac{\log T}{\frac{t\epsilon}{2K}}}+\frac{2C'}{t}$. Define "good" event $G$ such that at every round $t\geq 24K\log T/\epsilon$, for every arm $a$, $\hat{\rho}^t(a) \in [\rho(a)-w^t,\rho(a)+w^t]$ is true. Then $G$ happens with probability at least $1-\frac{2(K+1)}{T}$. The regret from rounds where $t<24K\log T/\epsilon$ can be bound by $24K\log T/\epsilon$, and henceforth we only focus on the rounds $t\geq 24K\log T/\epsilon$ where event $G$ is defined.

Let $a=1$ be the optimal arm. When $G$ is true and $t\geq 24K\log T/\epsilon$, the regret from this round is bound by:

\begin{equation}
    \begin{split}
        R^t &= \rho(1) \mu(1) - \rho(a^t)\mu(a^t) \\
            &\leq (\hat{\rho}(1)+w^t)\mu(1) -  (\hat{\rho}(a^t)-w^t)\mu(a^t)\\
            &\leq w^t(\mu(1)+\mu(a^t))\\
            &\leq 2w^t
    \end{split}
\end{equation}

The second inequality use the fact that $\hat{\rho}(1)\mu(1) \leq \hat{\rho}(a^t)\mu(a^t)$. Summing it over $t$ gives an upper bound on the regret from exploit rounds when $G$ is true and $t>24K\log T/\epsilon$: $$\textsc{Reg}_{\exploit | G }(T) \leq \sum_{t=1}^T 2w^t \leq 2(\sqrt{\frac{2KT\log T}{\epsilon}}+2C'\log T).$$

When event $G$ doesn't occur, $R_{\exploit | \overline{G} }(T)$ is at most $T$. 
We get
\begin{equation}
    \begin{split}
        \textsc{Reg}_{\exploit}(T) &\leq 24K\log T/\epsilon + \Pr\{G\}\cdot (2(\sqrt{\frac{2KT\log T}{\epsilon}})+2C'\log T) + \Pr\{\overline{G}\} T\\
        &\leq 24K\log T/\epsilon + 2(\sqrt{\frac{2KT\log T}{\epsilon}})+2C'\log T) + 2(K+1).
    \end{split}
\end{equation}

Adding $\textsc{Reg}_{\explore}(T)$ and $\textsc{Reg}_{\exploit}(T)$, we get 

$$\textsc{Reg}(T)\leq T\epsilon + 24K\log T/\epsilon + 2\sqrt{\frac{2KT\log T}{\epsilon}}+4C'\log T + 2(K+1).$$

By taking $\epsilon = 2(K\log T)^{1/3}T^{-1/3}$, the welfare regret is bound by $$\textsc{Reg}(T)\leq 4(K\log T)^{1/3}T^{2/3} + 4C'\log T + 12K^{2/3}T^{1/3}(\log T)^{4/3} + 2(K+1).$$

\paragraph{Proof for theorem \ref{thm:ccontextregret}}

First we give the guarantee on how many rounds are decided to be explore rounds, and how much corruption have been received in explore rounds. The expectation of the number of explore round $t_e$ at round $t$ is $\mathbbm{E}[t_e]=t\epsilon/K$. By Chernoff lower tail, taking $\delta=1/2$, then we have $\Pr\{t_e<\frac{t\epsilon}{2}\}\leq 1/T^2$ when $t\geq 24\log T/\epsilon$. Denote $C^t$ as the amount of corruption that received in the explore rounds at round $t$. 

The expectation of $C^t$ is no greater than $\epsilon C$. If $C<6\log T/\epsilon$, by Chernoff lower tail, taking $\delta = \frac{6\log T}{C \epsilon}$, we have $\Pr\{C^t > \epsilon(C+\frac{6\log T}{\epsilon})\} \leq 1/T^2$; if $C>6\log T/\epsilon$, take $\delta = \sqrt{\frac{6\log T}{C\epsilon}}$, we have $\Pr\{C^t >2\epsilon C\}\leq 1/T^2$. Combing above, when $t\geq 24\log T/\epsilon$, with probability at least $1-2/T$, we have $t_e>\frac{t\epsilon}{2}$ and $C^t<(C+\max\{C,\frac{6\log T}{\epsilon}\})\epsilon$ for all $j \in [K]$ and $t \in [T]$. Denote $C'=C+\max\{C,\frac{6K\log T}{\epsilon}\}$, then $C^t<C'\epsilon$.

Next we give the guarantee on the accuracy of accuracy on click probability of an agent $a$ induced by an expert $h$ $\Pi(a,h)$. Denote $r^t(a,h) = \mathbbm{1}(h(x^t)=a)\mathbbm{1}(a^t=a)\cdot c^t$. We have $\mathbbm{E}[r^t(a,h)]=\frac{\Pi(a,h)}{K}$ and $\hat{\Pi}^t(a,h)=K\frac{\sum_{t\in t_e}r^t(a,h)}{|t_e|}$.  The deviation from $\sum_{t\in t_e}r^t(a,h)$ to its expectation is given by the nature of stochastic, which can be bound by concentration inequalities, and the corruption being applied, which is bound by $C^t<C'\epsilon$. By Hoeffding inequality and union bound, for any expert $h$, any arm $i$, at any round $t\geq 24\log T/\epsilon$, with probability at least $2/T$,  we have $$|\hat{\Pi}^t(a,h)-\Pi(a,h)| \leq K\sqrt{\frac{2\log (T\cdot m\cdot K)}{t\epsilon}} + \frac{2C'}{t}.$$

Next we give the guarantee on when we can assume that $\alpha$-rational agents will bid truthfully. Define a good event $G$ that for any expert $h$, any arm $a$, at any round $t\geq 24\log(T)/\epsilon$, the following is true $$\Delta^t_C := |\hat{\Pi}^t(a,h)-\Pi(a.h)| \leq K\sqrt{\frac{2\log (T\cdot m\cdot K)}{t\epsilon}}+\frac{2C'}{t}.$$

Let $\mu'^t(a)$ be the bid that is actually the best bidding for agent $a$, we have: $$(\mu(a)-p^t(a,\mu(a)))\hat{g}^t(a,\mu(a))>(\mu_i-p^t(a,\mu'^t(a)))\hat{g}^t(a,\mu'^t(a)),$$ so the utility gap between bidding truthfully and best bidding $\Delta^t_u$ satisfies 
\begin{align*}
    \Delta^t_u &= (\mu(a)-p^t(a,\mu'^t(a)))g^t(a,\mu'^t(a)) - (\mu(a)-p^t(a,\mu(a)))g^t(a,\mu(a)) \\
    &\leq (\mu(a)-p^t(a,\mu'^t(a)))(\hat{g}^t(a,\mu'^t(a))+\Delta^t_c) - (\mu(a)-p^t(a,\mu(a)))(\hat{g}^t(a,\mu(a))-\Delta^t_c)\\
    &\leq [(\mu(a)-p^t(a,\mu'^t(a)))\hat{g}^t(a,\mu'^t(a)) - (\mu(a)-p^t(a,\mu(a)))\hat{g}^t(a,\mu(a))] +\\
    &(\mu(a)-p^t(a,\mu'^t(a))+\mu(a)-p^t(a,\mu(a))) \Delta^t_c\\
    &\leq 2 \Delta^t_C \\
\end{align*}

With probability at least $1-3/T$, for any $t\geq 24\log(T)/\epsilon$, $G$ is true and we have $$\Delta^t_u \leq 2K\sqrt{\frac{2\log (T\cdot m\cdot K)}{t\epsilon}}+\frac{2C'}{t}.$$ Therefore, when $G$ is true, $\Delta^t_u \leq \alpha$ is true for all $t\geq \max\{24\log(T)/\epsilon, \frac{8K^2 \log(T\cdot m\cdot K)}{\epsilon \alpha^2}\}+\frac{16C'}{\alpha}$. Considering the fact that $\alpha \leq 1$ and $K \geq 2$, $\max\{24\log(T)/\epsilon, \frac{8K^2 \log(T\cdot m\cdot K)}{\epsilon \alpha^2}+\frac{16C'}{\alpha}\}=\frac{8K^2 \log(T\cdot m\cdot K)}{\epsilon \alpha^2}+\frac{16C'}{\alpha}$. So with probability at least $1-3/T$, an $\alpha$-truthful agent will bid truthfully for any round $t \geq \frac{8K^2 \log(T\cdot m\cdot K)}{\epsilon \alpha^2}+\frac{16C'}{\alpha}$.

At last we can bound the regret. Define a good event $G'$ as all $\alpha$-truthful agent will bid truthfully when $t\geq \frac{8K^2 \log(T\cdot m\cdot K)}{\epsilon \alpha^2} +\frac{16C'}{\alpha} := t^*$. The probability that $G'$ is true is at least $1-3K/T$. The regret from the cases where $G'$ is not true is bound by $3K$. The regret from the rounds where agents might not be truthful is therefore bound by $t^*$. The regret from the explore rounds is bound by $T\epsilon$ in expectation. Next we focus on the regret from the exploit rounds after $t\geq t^*$. 

The empirical estimation on expert reward function is given by $\hat{R}^t(h) = \sum_{a \in \mathcal{A}} b^t(a) \hat{\Pi}(a,h)$. So the difference between the empirical estimation and true reward function of expert $h$ satisfies: $$\Delta^t_R = |\hat{R}^t(h,\bm{b}^t) - R(h,\bm{b}^t)| \leq \sum_{a \in \mathcal{A}} b^t(a) \Delta^t_C \leq K\Delta^t_C.$$

Then the difference between the expected reward of the best expert $h^{*}$ and the empirical best expert $h^t$ satisfies: 
\begin{align*}
    R(h^*,\bm{b}^t)-R(h^t,\bm{b}^t) &\leq (\hat{R}(h^*,\bm{b}^t)+\Delta^t_R) - (\hat{R}(h^t,\bm{b}^t)+\Delta^t_R)\\
    &= (\hat{R}(h^*,\bm{b}^t)-\hat{R}(h^t,\bm{b}^t)) + 2\Delta^t_R\\
    &\leq 2\Delta^t_R\\
    &\leq 2K \Delta^t_C\\
    &\leq 2K^2\sqrt{\frac{2\log (T\cdot m\cdot K)}{t\epsilon}}+\frac{4KC'}{t}
\end{align*}

So the total regret from these exploit rounds is bound by $$R_{\explore|t\geq t^*}(T) \leq \sum_{t=t^*}^T 2K^2\sqrt{\frac{2\log (T\cdot m\cdot K)}{t\epsilon}} +\frac{4KC'}{t} \leq 2K^2\sqrt{\frac{2T\log (T\cdot m\cdot K)}{\epsilon}} +4KC'\log T.$$

In total, the regret is bound by $$\textsc{Reg}(T)\leq T\epsilon+K^2\sqrt{\frac{8T\log (T\cdot m\cdot K)}{\epsilon}}+3K+\frac{8K^2 \log(T\cdot m\cdot K)}{\epsilon \alpha^2} + \frac{16C'}{\alpha}+4KC'\log T.$$

By taking $C'=0$ we can recover the results of lemma \ref{lem:error}, theorem \ref{thm:rationalguarantee}, and theorem \ref{thm:contextregret}.

\subsection{Example}\label{sec:example}

\textbf{Example}: Suppose there are three contexts $x_1, x_2, x_3$, and the probability of each to show up is $1/3$. There are three experts $h_1, h_2, h_3$ and three arms $a_1, a_2, a_3$. Let $\rho$ be the $3$ by $3$ matrix where $\rho_{i,j}$ denote the click through rate of arm $a_j$ when the context is $x_i$. Let \[
\rho = 
\begin{bmatrix}
    0.7 & 0.4 & 0.9\\
    0.2 & 0.2 & 0.5\\
    0.6 & 0.8 & 0.3
\end{bmatrix}.
\]

Let $D$ be a $3$ by $3$ matrix where $D_{i,j}$ denotes the arm that expert $h_i$ will suggest if the context is $x_j$. Let \[
D = 
\begin{bmatrix}
    a_0 & a_1 & a_2\\
    a_1 & a_2 & a_0\\
    a_2 & a_0 & a_1
\end{bmatrix}.
\]

\begin{figure}[htp]
    \centering
    \includegraphics[width=0.9\textwidth]{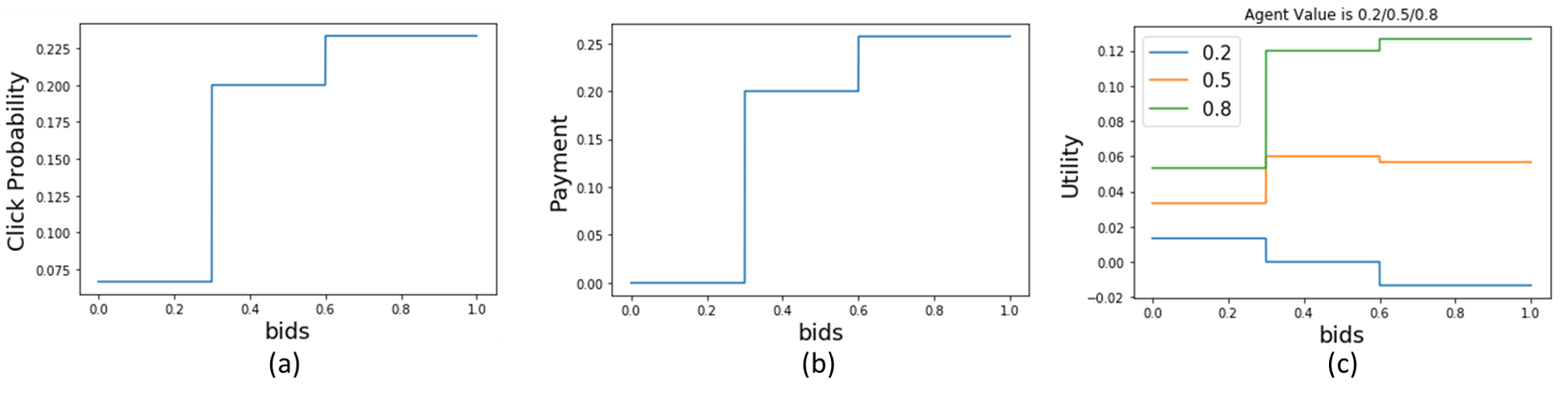}

    \caption{(a) The click probability of agent $a_0$ versus its bid (b) The truthful payment rule for agent $a_0$ versus its bid (c) The utility agent $a$ could get versus its bid. Each line represent a different agent value.}
        \label{fig:1}
\end{figure}

\begin{figure}[htp]
    \centering
    \includegraphics[width=0.9\textwidth]{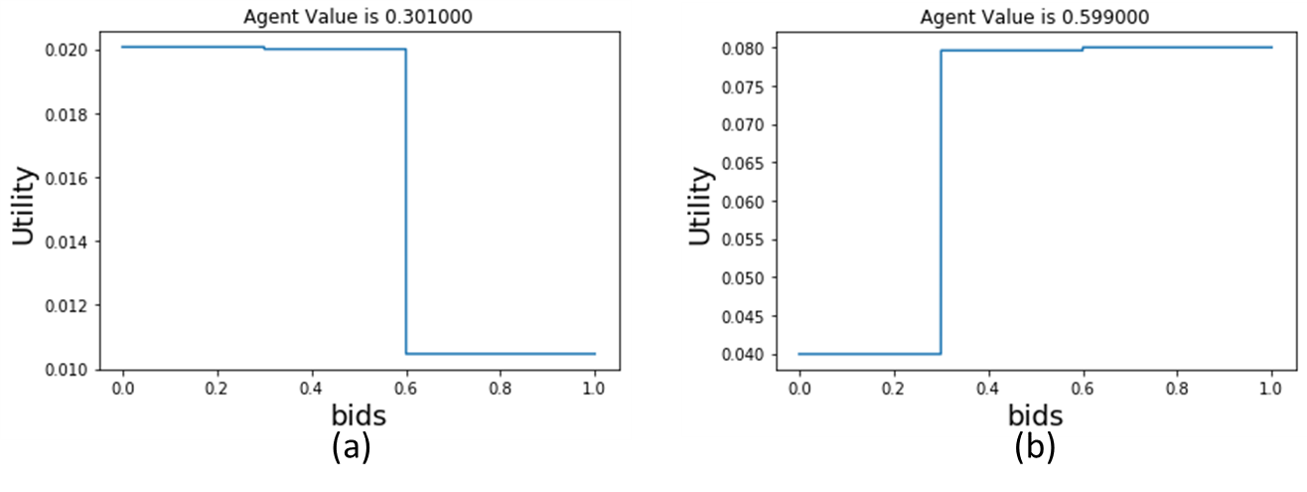}
   
    \caption{Given a payment rule which is slightly different from the truthful one, (a) is the utility for agent $a_0$ when its value is $0.301$, (b) is the utility for agent $a_1$ when its value is $0.599$.}
     \label{fig:2}
\end{figure}

Denote $\bm{b}$ be the bid profiles of all arms, and let $b(a_1) = 0.1$, $b(a_2) = 0.2$. Then the click probability for agent $a_0$ can be calculated by definition \ref{def:g}, which is plotted in figure \ref{fig:1} (a). The truthful payment rule for agent $a_0$ can be calculated by the formula given in lemma \ref{lem:payment}, which is plotted in figure \ref{fig:1} (b). The utility of agent $a_0$ can be calculated through the utility function $u^t(a,b) = (\mu(a)-p^t(a,b))g^t(a,b)$ for a given value $\mu(a)$. In figure \ref{fig:1} (c) we show the utility function for agent $a_0$ with different value, and we can find that the maximal of the utility is always achieved at the value, which means truthful bidding is the best strategy for $a_0$. However, if the payment rule is slightly deviated from the truthful one, then $a_1$ with some value will bid far from its value to optimize its utility. Change the payment rule by a little amount, more specifically, the payment is increased by $0.001$ when the bid is between $0.3$ and $0.6$, and decreased by $0.001$ when the bid is above $0.6$. In figure \ref{fig:2} (a), (b), we show that utility of $a_1$ with value $0.301$ and $0.599$ is not maximized at the corresponding value. To maximize the utility, $a_1$ could bid $0$ instead of $0.301$ and $1$ instead of $0.599$. Even though the optimal bid could be far from the true value, the extra utility agent $a_0$ can have by deviating from truthful bidding is as little as $0.0004$ and $0.00006$ in both cases respectively.

\subsection{When agents can observe contexts}\label{sec:observe}
When agents can observe the contexts, their don't optimize over the randomness of the context but can optimize for the specific realized context. More specifically, the click probability of an agent induced by an expert $g^t(\cdot)$ is no longer an expectation over context and depend on context, that is, the context $x$ is also a parameter in $g^t(a,b,x)$. Formally, we redefine the probability of an agent when agents can observe the context.  

\begin{definition}[Click probability of an agent (when agents can observe context)]
\label{def:g2}
Let $A$ be the allocation rule used by the mechanism, $H^t$ be the history till round $t$, and $\mathcal{H}$ be the experts class. For an agent $a$, fixing the bids of the other agents as $\bm{b}^t_{-a}$, then the probability that the agent $a$ get clicked in round $t$ by bidding $\bm{b}(a)=b$ is given by 

 $$g^t(a,b,x) = \Pr\{A(\mathcal{H}, H^t, x, b, \bm{b}^t_{-a})=a\}\rho(a,x)$$ 
 
 where $\rho(a,x)$ is the click through rate for agent $a$ on context $x$.
\end{definition}

In this case, the payment rule may also depend on the context $x$ and the utility of agents $u^t(a,b,x)$ now becomes $u^t(a,b,x) = (\mu(a)-p^t(a,b,x))g^t(a,b,x)$. Note that the equivalent for Lemma \ref{lem:payment} will also depend on the context $x$. To achieve truthfulness, now the mechanism needs to ensure that agents' utility is maximized by bidding truthfully for every individual context, which would require learning over the context space $\mathcal{X}$. Learning over $\mathcal{X}$ is not tractable and is exactly the reason we utilize the information from the context through experts. Next we show that even if the mechanism knows the expected reported welfare $R(h, \bm{b})$ for every expert $ h \in \mathcal{H}$ and bid profile $\bm{b}$, the greedy mechanism that follows the expert with maximum reported welfare cannot maintain truthfulness.

\begin{theorem}
If a mechanism always follows the expert $h^t = \argmax_h R(h,\bm{b}^t)$, then there is no payment rule to make the mechanism truthful. In other words, this greedy mechanism which recovers the highest expected welfare assuming every agent behaves truthfully cannot actually be truthful.
\end{theorem}

We can always find an instance with an appropriate expert class where $g^t(a,b,x)$ is not monotone on $b$ for some specific contexts $x$, and by the equivalent of Lemma \ref{lem:payment}, there is no payment rule to make such mechanism truthful.

Although contexts create incentive issues in this setting, with some strong assumptions on the experts class $\mathcal{H}$, truthfulness can still be achieved. We give such class of experts where we can maintain truthfulness for the greedy mechanism even if the agents can observe the contexts.

\begin{definition}[pre-ordered experts class]
An experts class $\mathcal{H}$ is called pre-ordered if the following holds. For any two expert $h_1,h_2 \in \mathcal{H}$ and any agent $a \in \mathcal{A}$, at least one of the following must be true

(1): $\mathbbm{1}\{h_1(x)=a\} \geq \mathbbm{1}\{h_2(x)=a\}$ is true for all $x \in \mathcal{X}$ 

(2): $\mathbbm{1}\{h_2(x)=a\} \geq \mathbbm{1}\{h_1(x)=a\}$ is true for all $x \in \mathcal{X}$.
\end{definition}

\textbf{Examples}:
Although pre-ordering in the experts is a strong assumption, there are some reasonable expert classes which are pre-ordered. One example is when experts are agents where each expert always suggest an agent regardless of context, i.e. the stochastic bandit mechanism setting. Another example is when there are only two arms, the context space is $\{x:x\in[0,1]\}$, and every expert $h$ is associated with a threshold $l(h)$ such that it suggests arm $1$ when $x \leq l(h)$ and arm $2$ when $x > l(h)$. In more general cases, this assumption doesn't hold. For example, even if the number of arms in the second example becomes $3$ and number of threshold becomes $2$ for each expert, then without any assumptions on the thresholds, this class of experts is not pre-ordered.

\begin{theorem}\label{thm:preoptimal}
If the expert class is pre-ordered, then there exists a payment rule such that the mechanism which always follows the expert $h^t = \argmax_h R(h,\bm{b}^t)$ is truthful. In other words, this greedy mechanism recovers the highest expected welfare, and can be implemented truthfully.
\end{theorem}

A pre-ordered expert class can guarantee that the probability of getting a click always increase when a bidder increases her bid, thus the \emph{click allocation} for each agent can be monotonically increasing, which makes it possible to have a payment rule that ensure truthfulness. 

\subsection{Stochastic $\epsilon$-greedy mechanism}

Here we present the stochastic $\epsilon$-greedy mechanism (Algorithm \ref{alg:robustmab}), which is basically a randomized version of the explore then commit mechanism. In each round, with a fixed probability $\epsilon$, the mechanism selects an arm uniformly at random, and these rounds are called \emph{explore rounds}. In this case each expert $h$ corresponds to a fixed arm, so the click probability of agent $a$ induced by expert $h$ $\Pi(a,h)$ is $\rho(a) = \mathbbm{E}_x[\rho(a,x)]$ for $a=h$ and $0$ for the other arms. Therefore we only need to estimate $\rho(a)$, the click through rate of each arm in this case, based on the results from the explore rounds only. The other others rounds are called \emph{exploit rounds} where the mechanism selects the empirically optimal arm given the current bids. The payment rule in the exploit rounds is a weighted second price rule where the weights are set to be the estimates on click through rates for arms, and the payment is always $0$ in explore rounds. In Algorithm \ref{alg:robustmab}, $\smax$ means the second highest.
{
\begin{algorithm}[htp]
    
  %\SetKwInOut{Input}{Input}
  \SetKwInOut{Param}{Parameters}
  \SetKwInOut{Init}{Initialize}

   \SetAlgoLined 
   \DontPrintSemicolon
 
%   \myalg{}{

  \Param{Number of arms $K$, Number of rounds $T$, exploration rate $\epsilon$}
  \Init{Set $\hat{\rho}^{0}_{j} \leftarrow 0$, $n^{0}_{j} \leftarrow 0$ for all $j \in [K]$
  }

  \For{$t=1, \ldots, T$}{
  
  Receive bid $b^t_j$ from arm $j$ for all $j \in [K]$

$\ell^t \leftarrow \begin{cases} 
      1 & \textnormal{w.p. } \epsilon \\
      0 & \textnormal{otherwise} \\
   \end{cases}$    \tcc*[r]{Explore or exploit}

  \eIf{$\ell^t = 1$}{
  \fbox{\tcc{\textbf{... Exploration Round ...}}}
  
    $a^t \leftarrow j $ uniformly at random for $j \in [K]$  \tcc*[r]{Select arm}
    
    Receive click result $c^t$
    
    Charge $p^t \leftarrow 0$  \tcc*[r]{Payment}

    \For(\tcc*[f]{Update empirical mean}){$j \in [K]$}{ 
        $ n^t_{j} \leftarrow n^{t-1}_{j} + {1}_{\{j=a^t\}}$ \\
         $ \displaystyle \hat{\rho}^{t}_{j} \leftarrow \frac{c^t\cdot {1}_{\{j=a^t\}}+\hat{\rho}^{t-1}_j\cdot n^{t-1}_{j}}{n^t_{j}}$
    }
  }{
  \fbox{\tcc{\textbf{... Exploitation Round ...}}}
    
    $a^t \leftarrow \argmax_j (\hat{\rho}^t_j \cdot b^t_j)$ \tcc*[r]{Select arm}
    Receive click result $c^t$
    
    Charge price $p^t \leftarrow \begin{cases} 
      \frac{\smax_j (\hat{\rho}^t_j \cdot b^t_j)}{\hat{\rho}^t_{a^t}} & \textnormal{\textbf{if} } c^t = 1 \\
      0 & \textnormal{otherwise} \\
   \end{cases}$ \tcc*[r]{Payment}
 }
}
\caption{Stochastic $\epsilon$-greedy Mechanism} \label{alg:robustmab}
\end{algorithm} 
}

\subsection{Auxiliary lemmas}

\begin{lemma}[Chernoff Bounds]
\label{lem:Chernoff}
Let $X_1,\ldots,X_n$ be independent random variables, and $X_i$ lies in the interval $[0,1]$. Define $X = \sum_{i=1}^n X_i/N$ and denote $E[X]=\mu$. For any $\delta \in [0,1]$, we have  \textbf{Chernoff lower tail:}  
$$Pr\{X<(1-\delta)\mu\} \leq \exp(-\frac{\mu \delta^2}{3})$$ and we have \textbf{Chernoff upper tail:}

$$Pr\{X>(1+\delta)\mu\}  \leq \begin{cases} 
     \exp(-\frac{\mu \delta}{3}) & \textnormal{\textbf{for} } \delta > 1 \\
      \exp(-\frac{\mu \delta^2}{3}) & \textnormal{\textbf{for} } \delta \in [0,1]\\
   \end{cases}$$
\end{lemma} 

\newpage

\end{document}